\documentclass[sigconf]{acmart}
\usepackage[utf8]{inputenc}
\usepackage[T1]{fontenc}
\usepackage{graphicx}
\usepackage{grffile}
\usepackage{longtable}
\usepackage{wrapfig}
\usepackage{rotating}
\usepackage[normalem]{ulem}
\usepackage{amsmath}
\usepackage{textcomp}
\usepackage{amssymb}
\usepackage{capt-of}
\usepackage{subcaption}

\usepackage{alltt}
\usepackage{amsthm}
\newtheorem{theorem}{Theorem}
\newtheorem{lemma}[theorem]{Lemma}
\usepackage[utf8]{inputenc}
\usepackage{verbatim}
\usepackage{fancyvrb}

\SaveVerb{carot}=^=

\CustomVerbatimCommand{\VerbatimInputNumbered}{VerbatimInput}
{ numbers=left }

\AtBeginDocument{%
  \providecommand\BibTeX{{%
    \normalfont B\kern-0.5em{\scshape i\kern-0.25em b}\kern-0.8em\TeX}}}

\setcopyright{acmcopyright}
\copyrightyear{2020}
\acmYear{2020}

\AtBeginDocument{%
  \providecommand\BibTeX{{%
    \normalfont B\kern-0.5em{\scshape i\kern-0.25em b}\kern-0.8em\TeX}}}

\copyrightyear{2020}
\acmYear{2020}
\setcopyright{acmlicensed}\acmConference[FODS '20]{Proceedings of the 2020 ACM-IMS Foundations of Data Science Conference}{October 19--20, 2020}{Virtual Event, USA}
\acmBooktitle{Proceedings of the 2020 ACM-IMS Foundations of Data Science Conference (FODS '20), October 19--20, 2020, Virtual Event, USA}
\acmPrice{15.00}
\acmDOI{10.1145/3412815.3416896}
\acmISBN{978-1-4503-8103-1/20/10}

\begin{document}
\fancyhead{}
\title{Transforming Probabilistic Programs for Model Checking}

\author{Ryan Bernstein}
\affiliation{%
  \institution{Columbia University}
  \city{New York, USA}
}

\author{Matthijs V\'ak\'ar}
\affiliation{\institution{Utrecht University}
\city{Utrecht, Netherlands}}

\author{Jeannette Wing}
\affiliation{\institution{Columbia University}
\city{New York, USA}}

\renewcommand{\shortauthors}{Bernstein, V\'ak\'ar, Wing}

\begin{abstract}
Probabilistic programming is perfectly suited to reliable and transparent data science, 
as it allows the user to specify their models in a high-level language without worrying about
the complexities of how to fit the models.
Static analysis of probabilistic programs presents even further opportunities for enabling a high-level style of programming, by
automating time-consuming and error-prone tasks. 
We apply static analysis to probabilistic programs to automate large parts of two crucial model checking methods:
Prior Predictive Checks and Simulation-Based Calibration.
Our method transforms a probabilistic program specifying a density function into an efficient forward-sampling form.
To achieve this transformation, we extract a factor graph from a probabilistic program using static analysis, generate a set of proposal directed acyclic graphs using a SAT solver, select a graph which will produce provably correct sampling code, then generate one or more sampling programs.
We allow minimal user interaction to broaden the scope of application beyond what is possible with static analysis alone.
We present an implementation targeting the popular Stan probabilistic programming language, automating large parts of a robust Bayesian workflow for a wide community of probabilistic programming users.\vspace{-3pt}
\end{abstract}





\maketitle
\vspace{-2pt}
\section{Introduction}
\vspace{-2pt}
\label{sec:orgb35e20e}
Probabilistic programming is widely used for data analysis across many fields, such as astrophysics \cite{lieu2017hierarchical}, (seasonal) forecasting \cite{taylor2018forecasting}, pharmacology \cite{becker2017therapeutic}, and public health \cite{burnett2018global}.
It is rapidly gaining in popularity\footnote{This can be
seen e.g. from download statistics of RStan on CRAN,
from the citation rate of \cite{carpenter2017stan}, 
and from the push by companies like Google, Facebook,
and Uber to each develop their own in-house probabilistic 
programming language.}.
The key idea is to let the user specify a probabilistic model as a high-level program, while the compiler solves the difficult task of generating performant code to fit the model using some complex inference algorithm.
Free from details of the inference implementation, probabilistic programs are the ideal medium for data science: replicable, interpretable, and communicable, while staying on the cutting edge of inferential power.

Representing our data science models as programs also affords us the opportunity to apply code analysis and transformation techniques from decades of computer science research.
In much the same way that software engineers have tools to check or generate code, data scientists could have tools to verify or construct their models.
This intersection of static analysis and probabilistic programming is a largely unexplored area that could make data science simultaneously more reliable, efficient and approachable.

In this paper, we apply code analysis and transformation techniques to automate large parts of two model checking techniques: Prior Predictive Checking (PPC) \cite{gabry2019visualization} and Simulation-based Calibration (SBC) \cite{SBC}.
PPC addresses the question, "Does the model as we have written it match our actual domain knowledge?"; SBC addresses the question "Can our inference algorithm fit our specified model reliably?".
Together, these checks make up a vital part of a robust Bayesian workflow \cite{kennedy2019experiment}, and they are used ubiquitously within the Bayesian probabilistic modeling community.

PPC and SBC both require random draws from the \emph{prior} and \emph{prior-predictive} distributions.
The prior is the distribution over model parameters prior to the data being considered, while the prior-predictive is the distribution over the data implied by the prior.

Currently, to apply PPC or SBC, users manually write sampling code which must correspond exactly to the prior and prior-predictive distributions of their particular model.
At best, this process is time consuming, requires expertise, and duplicates the information and maintenance cost of the original program.
At worst, the produced prior-predictive distribution is wrong, causing PPC and SBC to produce misleading results, which defeats the purpose of these quality assurance methods. 

To mitigate this issue, we present a method to transform a probabilistic program into one which efficiently draws samples from the prior and prior-predictive distributions of the model specified by the original program.
Our process either produces a guaranteed-correct and efficient program or, in rare cases, rejects the task as impossible, all with minimal user involvement.
We focus on the Stan language for our implementation and primary reference point, as Stan is one of the most widely used probabilistic programming languages, as measured by package downloads and citations \cite{carpenter2017stan}.

Central in our approach is a novel translation from a factor graph -- a probabilistic graphical model giving a convenient intermediate representation of a probabilistic program -- into a directed acyclic graphical (DAG) model, which admits efficient forward sampling.

This paper makes the following core contributions:
\begin{enumerate}
\item It presents a technique for automatically computing all possible DAGs which represent the same conditional independence information as a given factor graph, together with a proof of its soundness and completeness.
\item It shows how to use this technique to  derive efficient  forward sampling code from a probabilistic program presented in a relatively unstructured representation, whenever possible.
\item It presents an implementation targeting the Stan probabilistic programming language, automating large parts of a robust Bayesian workflow for a wide community of probabilistic programming users.
\end{enumerate}

The source code of our implementation is available online \cite{implementation}.
\section{Background}
\label{sec:org76a8ce4}
\subsection{Probabilistic programs}
\label{sec:org590b560}
Probabilistic models represent a joint probability distribution over a set of unobserved parameters and observed data.
One representation of a probabilistic model is as a sequence of parameters drawn from individual distributions.
For example:
 \begin{align*}
   &\mu \sim Normal(0, 1);\quad
   \sigma \sim LogNormal(1, 3);\quad
   x \sim Normal(\mu, \sigma);
\end{align*}

This represents a probabilistic model in which \(\mu\), \(\sigma\) and \(x\) are drawn from distributions.
We can efficiently draw samples of \(\mu\), \(\sigma\) and \(x\) with the following process:
\begin{enumerate}
\item draw from a \(Normal(0,1)\) distribution; bind the result to the variable \(\mu\).
\item draw from \(LogNormal(1,3)\); bind the result to \(\sigma\).
\item draw from \(Normal(\mu,\sigma)\); bind the result to \(x\).
\end{enumerate}
We call this a \emph{forward sampling representation} of the probabilistic model: each variable can be drawn individually in sequence.

However, suppose instead that we want to draw
\(\mu\) and \(\sigma\) but the variable \(x\) is observed.
We then need to perform probabilistic inference to determine the \emph{posterior} distributions for \(\mu\) and \(\sigma\) given the observed \(x\),
and we should draw \(\mu\) and \(\sigma\) from this posterior.

Probabilistic programs represent probabilistic models, such as the example above, in a way that facilitates inference algorithms to draw from the posterior distributions of the unobserved parameters.

In the Stan probabilistic programming language, models are represented in terms of the joint density function over all observed and unobserved variables.
This function of the variables, called the \emph{log-probability density function} or \emph{lpdf}, is then typically passed to a Markov-chain Monte Carlo (MCMC) algorithm which provides draws from the posterior distribution \cite{MCMC}.
We could represent the example probabilistic model as the following Stan program:
\begin{alltt}
data \{
  real x;
\}
parameters \{
  real mu;
  real sigma;
\}
model \{
  target += normal\(\_\)lpdf(mu | 0, 1);
  target += lognormal\(\_\)lpdf(sigma | 1, 3);
  target += normal\(\_\)lpdf(x | mu, sigma);
\}
\end{alltt}
This program defines a joint \texttt{lpdf} function of the (observed) data \texttt{x} and the (unobserved) model parameters \texttt{mu} and \texttt{sigma}.
The special variable \texttt{target} holds the value of the \texttt{lpdf} and is implicitly initialized to \texttt{0} and returned at the end of \texttt{model}.
Each variable contributes to \texttt{target} according to its distribution's density, and the final joint density is the product (or sum, in log space) of these~contributions.

This \emph{density representation} of the program allows for powerful posterior inference capabilities, but it is no longer straightforward to follow the efficient forward-sampling procedure.
 The challenge we address in this paper is to allow the user to write their program once as a density representation, and to transform it automatically 
 to its corresponding a forward-sampling representation.
\subsection{Efficiently sampling from model parameters}
\label{sec:org5a66153}

In the modeling process, we sometimes want to sample directly from our model of the variables, for example to perform PPC and~SBC.

A practical barrier to using tools like SBC is the time it takes to naively sample a large number of parameters from the model, when it is presented in a density representation.
In our example, the sampling process may be written as:
\begin{alltt}
(mu, sigma, x) \(\sim\) sample(normal\(\_\)lpdf(mu | 0, 1)
                        \(\texttt{+}\) lognormal\(\_\)lpdf(sigma | 1, 3)
                        \(\texttt{+}\) normal\(\_\)lpdf(x | mu, sigma));
\end{alltt}
Here, \texttt{sample} is some sampling algorithm, such as an MCMC method, which takes an \texttt{lpdf} function of the distribution of the variables and produces draws.
Running this algorithm is not an especially efficient way to sample.
Instead, the following would be ideal:
\begin{alltt}
mu \(\sim\) sample(normal\(\_\)lpdf(mu | 0, 1));
sigma \(\sim\) sample(lognormal\(\_\)lpdf(sigma | 1, 3));
x \(\sim\) sample(normal\(\_\)lpdf(x | mu, sigma));
\end{alltt}
This sampling method is potentially faster in two ways:
\begin{enumerate}
\item It decomposes the sampling process to draw each variable individually.
This scales and parallelizes easily.
\item It isolates the distributions of each variable, allowing us to potentially recognize common distributions with known efficient sampling algorithms, such as the Normal distribution.
\end{enumerate}
This sequential approach, called forward sampling, is often significantly faster than sampling from the program as a whole using an MCMC method, and has better convergence properties \cite{MCMC}.
\subsection{Translation challenges}
\label{sec:orgac3c72d}
  Ideally, we would like to translate programs written in a general \texttt{lpdf}-function form into a form amenable to forward sampling. 
There are a number of issues that make this translation challenging:

\begin{enumerate}
\item Programmers are free to write the \texttt{lpdf} function as unconstrained code, with intermediate variables and program structures that muddle the dependencies between the variables.
\item There may be more than one way to produce a valid translation.
For example, consider the following \texttt{lpdf} function:
\begin{alltt}
parameters \{
  real x, y;
\}
model \{
   target += f1(x);
   target += f2(y);
   target += f3(x, y);
\}
\end{alltt}
There are at least two valid interpretations of this \texttt{lpdf} program as sequential sampling statements:
\begin{alltt}
x \(\sim\) sample(f1(x))
y \(\sim\) sample(f2(y) + f3(x, y))
\end{alltt}
and
\begin{alltt}
y \(\sim\) sample(f2(y))
x \(\sim\) sample(f1(x) + f3(x, y))
\end{alltt}
There is not enough information in the original program to determine which of these distinct sampling distribution assignments is correct.
\item There are some \texttt{lpdf} functions which cannot be decomposed into a sequence of sampling statements:
\begin{alltt}
parameters \{
  real x, y, z;
\}
model \{
  target += f1(x, y)
  target += f2(x, z)
  target += f2(y, z)
\}
\end{alltt}
Which parameter should be drawn first? 
As every parameter depends on another,
this represents a directed graphical model with a cycle. Hence, forward sampling is not possible.
\end{enumerate}
\subsection{Our approach}
\label{sec:org809a50e}

We present a process for translating probabilistic programs in \texttt{lpdf} form into forward-sampling programs.
We start by observing that, at compile-time, the conditional independence structure of an \texttt{lpdf}-function probabilistic program can be distilled into a factor graph, which is an undirected graph of factors (such as \texttt{f1} and \texttt{f2} in the examples above) and variables (\texttt{x}, \texttt{y}, and \texttt{z}).
Further, we represent probabilistic models which can be forward sampled by the more informative abstraction of a directed acyclic graphical model.
Using these abstractions as intermediates between probabilistic programs and forward-sampling programs, our approach works as follows:
\begin{enumerate}
\item Extract a factor graph from the \texttt{lpdf}-function program by applying dependency analysis to the program source code.
\item Perform a graph transformation from the factor graph to a directed acyclic graph (DAG), when possible.
We describe this process in the methods section.
\item Produce forward-sampling code for the variables in topological order of the DAG.
\end{enumerate}

\section{Methods}
\label{sec:org2991b5c}
\label{sec:methods}
In this section, we present our procedure for translating a factor graph derived from a probabilistic program into to DAG from which we can efficiently sample.
We define the necessary terms below, then define the two graph abstractions, present a sequence of algorithms, and prove soundness and completeness properties of the methods.

For an input probabilistic program \(P\), let \(S(P)\) be the set of statements in \(P\) and let \(V(P)\) be the set of identifiers for variables in~\(P\).

Intuitively, a \emph{factor} is a fragment of the program \(P\) that directly contributes to the joint density that \(P\) represents.
To be precise, a factor \(f\) is a statement \(s\in S(P)\) along with the set of statements on which \(s\) depends.
Let \(F(P)\) be the set of all factors in a program, then \(f \in F(P) \subseteq S(P) \times \mathcal{P}(S(P))\), where $\mathcal{P}$ indicates the powerset.

When we write \(f(v_1, v_2,  \dots )\) with \(v_1, v_2,  \dots \in V(P)\), we interpret \(f\) as the function which corresponds to the density contribution calculated by \(f\), given specified values for  \(v_i\).

Note that throughout section \ref{sec:methods}, we will work with \(pdf\) functions rather than \(log\) \(pdf\) functions, despite Stan factors being in \(lpdf\) form.
We find that it is more intuitive to reason about densities outside of \(log\)-space.
All of the results in this section can mechanically be translated to \(log\)-space, if desired.

\subsection{Factor Graphs}
\label{sec:org6441d27}
\label{sec:fg}

A \emph{factor graph} represents a probability distribution as a product of factors that make up the distribution's density function.
A factor graph \(G_{Fac}\) is an undirected, bipartite graph with two sets of vertices: the sets of variables \(V(G_{Fac})\) and factors \(F(G_{Fac})\).
For a program \(P\), we use 
\(V(G_{Fac})=V(P)\) and \(F(G_{Fac})=F(P)\).
The edge set \(E(G_{Fac}) \subseteq V(G_{Fac}) \times F(G_{Fac})\) connects each factor to the variables that it depends on. Let \(Nei(G_{Fac}, f) = \{v \mid (v, f) \in E(G_{Fac})\}\) be the set of variables which are \(f\)'s neighbors in \(G_{Fac}\).
A factor graph then defines a joint density function:
$$pdf_{G_{Fac}}(V(G_{Fac})) = \prod_{f \in F(G_{Fac})} f(Nei(G_{Fac}, f))$$

\subsection{Directed Acyclic Probabilistic Graphical Models}
\label{sec:orgc898211}
A \emph{directed acyclic probabilistic graphical model} (DAG) represents a joint probability distribution as the product of conditional density functions which are associated with each variable.
A DAG \(G_{DAG}\) consists of sets \(V(G_{DAG})\) of vertices 
and 
\(E(G_{DAG})\subseteq V(G_{DAG}) \times V(G_{DAG})\)
of directed edges,
together with, for each \(v\in V(G_{DAG})\), a 
real-valued function \(D_v\).
We think of \(D_v\) as an (unnormalized)
conditional density
for \(v\) given its parents in the graph.
The graph then represents a joint density over the variables as:
$$pdf_{G_{DAG}}(V(G_{DAG})) = \prod_{v \in V(G_{DAG})} D_v(v \mid Par(G_{DAG}, v))$$
For a program \(P\), we can choose \(V(G_{DAG}) = V(P)\) 
and 
\(E(G_{DAG}) \subseteq V(G_{DAG}) \times V(G_{DAG})\).
Given a function \(A_{G_{DAG}} : V(G_{DAG}) \rightarrow \mathcal{P}(F(P))\) to assign to each variable the factors that make up its conditional density function, we can define 
$$D_v(v \mid Par(G_{DAG}, v)) = \prod_{f\in A_{G_{DAG}}(v)} f(Par(G_{DAG}, v)).$$

We will refer to variables without any parents in a DAG as root variables and variables without any children in a DAG as leaf variables.
Let \(Par(G_{DAG}, v) = \{v' \mid (v', v) \in E(G_{DAG})\}\) be the set of parents of \(v\) in \(G_{DAG}\).

We call a density   \(D_v(v\mid v_1, \dots, v_N)\) \emph{constant-normalized} if its total mass is constant with respect to the variables being conditioned on:
$$\forall i\in [1, \dots, N],\ \ \frac{\partial}{\partial v_i} \Big( \int D_v(v \mid v_1, \dots, v_N)\ dv \Big) = 0.$$

Let a DAG be called a \emph{constant-normalized DAG} if all \(D_v\) are constant-normalized.
We note that the typical definition of a DAG requires that each \(D_v\) be a conditional probability distribution, and so it would trivially fit what we call a constant-normalized DAG.

\subsubsection{Sampling from constant-normalized DAGs}
\label{sec:org1e36cd2}
\label{sec:sampling}
We show that sampling from a constant-normalized DAG \(G\) in topological order is consistent with the joint density of the graph.

Let \texttt{sample} be a procedure for sampling a value from a (not necessarily normalized) density function \(pdf\):
$$P(v \leftarrow \texttt{sample}(pdf)) \propto pdf(v)$$


We define a \emph{conditioned DAG} \(G\big|_{v\leftarrow \bar{v}}\) for some DAG \(G\) and root variable \(v\in V(G)\) to be a DAG with variables \(V(G\big|_{v\leftarrow \bar{v}}) = V(G) - v\), edges \(E(G\big|_{v\leftarrow \bar{v}}) = \{(v', v'') \mid (v', v'')\in E(G), v'\neq v\}\), and conditional densities \(D_{v'}|_{v\leftarrow \bar{v}}(v'|Par(v')) = D_{v'}(v'|Par(v'),\bar{v})\).

Let \(N=|V(G)|\) be the number of variables in \(G\), and let \(v_1, \dots, v_N\) be any topological order of the variables in \(G\), so that \(v_1\) is a root and \(v_N\) is a leaf.
Let \(V_i\) represent the set of variables \(\{v_1, \dots, v_i\}\), and \(\bar{V}_i\) represent a set of draws \(\{\bar{v}_1, \dots, \bar{v}_i\}\) for these variables.
Then, we iterate the previous construction and write 
\(G\big|_{V_0\leftarrow\bar{V}_0}=G\) and 
\(G\big|_{V_i\leftarrow\bar{V}_i}=G\big|_{V_{i-1}\leftarrow \bar{V}_{i-1}}\big|_{v_i\leftarrow \bar{v}_i}\).

Let \(P_G(v)\), \(P_G(v_1,\dots v_N)\), and \(P_G(v|\dots)\) be marginal, joint and conditional distributions of the variables under the joint distribution of \(G\) defined by the joint density \(pdf_{G_{DAG}}(V(G))\).

\begin{lemma}
\label{lm:marginal}
\noindent For any root variable \(v_r\) in a constant-normalized DAG \(G\), \(D_{v_r}(v_r)\propto P_G(v_r)\).
\end{lemma}

\begin{proof}
Let \(v_\ell\) be any leaf variable in \(G\) such that \(v_\ell \neq v_r\):
    \begin{align*}
  P_G(v_r) &= \int_{V(G)\setminus v_r} pdf_{G_{DAG}}(V(G)) \prod_{v\in V(G)\setminus v_r}dv \\
  &= D_{v_r}(v_r)\int_{V(G)\setminus v_r} \prod_{v\in V(G)\setminus v_r} D_v(v\mid Par(v)) dv \\
  &= D_{v_r}(v_r)\int_{V(G)\setminus v_r\setminus v_\ell} \Big( \int_{v_\ell} D_{v_\ell}(v_\ell\mid Par(v_\ell)) dv_\ell \Big)\\ &\qquad\prod_{v\in V(G)\setminus v_r \setminus v_\ell} D_v(v\mid Par(v)) dv \\
  &= \Big( \int_{v_\ell} D_{v_\ell}(v_\ell\mid Par(v_\ell)) dv_\ell \Big) D_{v_r}(v_r)\\&\qquad \int_{V(G)\setminus v_r\setminus v_\ell}\prod_{v\in V(G)\setminus v_r \setminus v_\ell} D_v(v\mid Par(v)) dv \\
  &\propto  D_{v_r}(v_r)\int_{V(G)\setminus v_r\setminus v_\ell} \prod_{v\in V(G)\setminus v_r \setminus v_\ell} D_v(v\mid Par(v)) dv \\
  &= P_{G\setminus v_\ell}(v_r).
\end{align*}
Here, the fourth equality follows by constant-normalization.

We then induct with a new \(v_\ell\) until there are no non-\(v_r\) leaf variables.
At that point we have \(P_G(v_r) \propto P_{G_{v_r}}(v_r)\), where \(G_{v_r}\) is \(G\) without any descendants of \(v_r\).
Since \(v_r\) is then both a root and a leaf variable in \(G_{v_r}\):
\begin{align*}
  P_G(v_r) &\propto P_{G_{v_r}}(v_r) \propto D_{v_r}(v_r)
\end{align*}
\end{proof}

\begin{lemma}
\label{lm:conditional}
\noindent For topologically ordered variables \(v_1, \dots, v_{i-1}\) in a constant-normalized DAG \(G\),
$$P_{G|_{V_{i-1}\leftarrow \bar{V}_{i-1}}}(v_i)=P_G(v_i \mid \bar{V}_{i-1})$$
\end{lemma}

\begin{proof}
\begin{align*}
  P_G&(v_i \mid \bar{V}_{i-1}) \\
  &= \int_{V(G)\setminus v_i \setminus V_{i-1}} \prod_{v\in {V(G) \setminus V_{i-1}}} D_v(v\mid Par(v)\setminus V_{i-1}, \bar{V}_{i-1}) dv \\
  &= \int_{V(G)\setminus v_i \setminus V_{i-1}} \prod_{v\in {V(G) \setminus V_{i-1}}} D_v\big|_{V_{i-1}\leftarrow \bar{V}_{i-1}}(v\mid Par(v)\setminus V_{i-1}) dv \\
  &= P_{G|_{V_{i-1} \leftarrow \bar{V}_{i-1}}}(v_i)
\end{align*}
\end{proof}

\begin{theorem}
\label{thm:sampling}
\noindent When each of the variables of \(G\) \(v_1, \dots, v_N\) are drawn in topological order according to \(\bar{v}_i\leftarrow \texttt{sample}(D_{v_i}\big|_{V_{i-1}\leftarrow \bar{V}_{i-1}})\), the draws \(\bar{V}_N\) are distributed according to \(P_G(\bar{V}_N)\).
\end{theorem}

\begin{proof} We calculate
\begin{align*}
  P(\bar{v}_1\leftarrow & \texttt{sample}(D_{v_1}), \dots, \bar{v}_N\leftarrow \texttt{sample}(D_{v_N}))\\
  = P(&\bar{v}_1\leftarrow \texttt{sample}(D_{v_1})) \cdot \\ 
    P(&\bar{v}_2\leftarrow \texttt{sample}(D_{v_2}) \mid \bar{v}_1 ) \cdot \\ 
  &\dots \\
  P(&\bar{v}_N\leftarrow \texttt{sample}(D_{v_N}) \mid \bar{V}_{N-1}).
\end{align*}

Since \(v_1\) is a root variable in \(G\), by the definition of \texttt{sample} and Lemma \ref{lm:marginal},
\(P(\bar{v}_1\leftarrow \texttt{sample}(D_{v_1})) \propto D_{v_1}(\bar{v}_1) \propto P_G(v_1)\),
so \(P(\bar{v}_1\leftarrow \texttt{sample}(D_{v_1})) = P_G(\bar{v}_1)\).
By the same reasoning, since each \(v_i\) is a root variable in \(G\big|_{V_{i-1}\leftarrow \bar{V}_{i-1}}\),
\(P(\bar{v}_i\leftarrow \texttt{sample}(D_{v_i}\big|_{V_{i-1}\leftarrow \bar{V}_{i-1}})) \propto
    D_{v_1}\big|_{V_{i-1}\leftarrow \bar{V}_{i-1}}(\bar{v}_1) \propto
    P_{G|_{V_{i-1}\leftarrow \bar{V}_{i-1}}}(v_i)\),
so \(
    P_{G|_{V_{i-1}\leftarrow \bar{V}_{i-1}}}(v_i)=\linebreak P(\bar{v}_1\leftarrow \texttt{sample}(D_{v_i}\big|_{V_{i-1}\leftarrow \bar{V}_{i-1}})) \).
Making those substitutions:
\begin{align*}
  P(\bar{v}_1\leftarrow & \texttt{sample}(D_{v_1}), \dots, \bar{v}_N\leftarrow \texttt{sample}(D_{v_N}))\\
  = P&_G(\bar{v}_1) \cdot \\ 
  P&_{G|_{v_1=\bar{v}_1}}(\bar{v}_2) \cdot \\ 
  &\dots \\
  P&_{G|_{V_{N-1}\leftarrow \bar{V}_{N-1}}}(\bar{v}_N). 
\end{align*}
Then by Lemma \ref{lm:conditional}, for each \(i\):
\(P_{G|_{V_{i-1}=\bar{V}_{i-1}}}(\bar{v}_i) 
    = P_G(\bar{v}_i | \bar{V}_{i-1})\).
\begin{align*}
  P(\bar{v}_1\leftarrow & \texttt{sample}(D_{v_1}), \dots, \bar{v}_N\leftarrow \texttt{sample}(D_{v_N}))\\
= P&_G(\bar{v}_1) \cdot \\ 
  P&_G(\bar{v}_2 | \bar{v}_1) \cdot \\ 
  &\dots \\
  P&_G(\bar{v}_N | \bar{V}_{N-1}) \\ 
  = P&_G(\bar{v}_1, \bar{v}_2, \dots, \bar{v}_N).
\end{align*}
\end{proof}

Constant-normalization is necessary for Lemma \ref{lm:marginal}.
If any \(D_v\) is not constant-normalized, the sampling process may not be consistent with the joint distribution.

We define, in pseudocode, \(\texttt{sampleG}(G)\), a procedure which draws a sample for each variable in a DAG \(G\) in topological order. Writing \(+\!\!\!+\) for the sequence concatenation operator,
\begin{align*}
  \texttt{sampleG}(G) =& \texttt{ let } v_r \in roots(G),\ \bar{v}_r \leftarrow \texttt{sample}(A_{G}(v_r)) \texttt{ in}\\
  &\ \texttt{[}\bar{v}_r\texttt{]} +\!\!\!+\ \texttt{sampleG}(G\big|_{v_r\leftarrow \bar{v}_r})
\end{align*}
We pass \texttt{sample} the set \(A_G(v_r)\) of assigned factors to allow for flexibility of implementation.
The graph is thus reduced in (any) topological order, with each variable replaced by a draw from its marginal distribution.

\subsection{A relation between Factor Graphs and DAGs}
\label{sec:org6d838db}
Factor graphs and DAGs are both representations of the joint density of a set of variables.
We say that a DAG \(G_{DAG}\) is a \emph{sound} transformation of a factor graph \(G_{Fac}\) if:
\begin{enumerate}
\item \(V(G_{DAG}) = V(G_{Fac})\)
\item \(pdf_{G_{DAG}}(V(G_{DAG})) = pdf_{G_{Fac}}(V(G_{Fac}))\)
\end{enumerate}

Since \(G_{DAG}\) is a DAG, it also must be acyclic.

Let the relation \(\sigma(G_{Fac}, G_{DAG})\) be true if and only if \(G_{DAG}\) is a sound transformation of \(G_{Fac}\).  
\subsection{Defining a transformation from Factor Graphs to DAGs}
\label{sec:org2395953}
Our strategy to transform factor graphs into DAGs will be to reinterpret each factor as part of the conditional density function of some variable.

A factor can only be part of the conditional density of a variable if that factor is a function of that variable, so factors will always be assigned to variables with which they share an edge in the factor graph.
The assignment of factors to conditional densities can thus be thought of as selecting a subset of edges from the factor graph, which we call the \emph{edge selection set}, \(s \subseteq E(G_{Fac})\).

To produce a DAG given a factor graph \(G_{Fac}\) and an edge selection set \(s\), we define a function \(G_{DAG} = C(G_{Fac}, s)\), which we call the \emph{contraction function}.
The contraction function produces a DAG with a vertex for each variable and directed edges built according to the edge selection set:
\begin{align*}
V(C(G_{Fac}, s)) =& V(G_{Fac})\\
E(C(G_{Fac}, s)) =&\\
\{(v_a, v_b) \mid & v_a, v_b\in V(G_{Fac}),\\
                    &\exists f \in F(G_{Fac}),\ (v_b, f) \in s\ \wedge\ (v_a, f) \in E(G_{Fac})\}\\
A_{C(G_{Fac}, s)}(v) =& \{ f \mid (v', f) \in s, v' = v \}
\end{align*}
The contraction function contracts all of the edges in the selection set, adding each variable which contributes to a contracted factor as a parent. 
The joint density of the resulting DAG then is
\begin{align*}
pdf_{G_{DAG}}(V(C(G_{Fac}, s))) &= \prod_v D_v(v \mid Par(C(G_{Fac},s), v))\\
 &= \prod_v \prod_{f\mid (v, f) \in s} f(Nei(G_{Fac}, f))
\end{align*}

\subsection{Computing sound transformations between Factor Graphs and DAGs}
\label{sec:org70e604b}
\label{sec:computing}

Our goal is to find the set \(S^*\) of edge selection sets \(s\) which produce a sound and constant-normalized DAG \(C(G_{Fac}, s)\).
Our strategy will be as follows:

\begin{enumerate}
\item \emph{Identify the set \(r\) of recognizable edges.}
Use heuristics to immediately identify some factors as constant-normalized conditional densities for a neighboring variable, producing the set of \emph{recognizable edges}, \(r\).
Discussed in section \ref{sec:recognizing}.
\item \emph{Construct the set \(S\) of valid edge selection sets.}
Construct the set \(S\) of edge selection sets \(s\) which produce a sound DAG \(C(G_{Fac}, s)\) such that \(r \subseteq s\).
We accomplish this by use of a SAT solver.
Discussed in section \ref{sec:computing}.
\item \emph{Query the user when necessary.}
When it is ambiguous whether a conditional density is constant-normalized, query the user for an assertion of constant-normalization.
Then, filter the set \(S\) down to the subset \(S^*\) of those edge selection sets consistent with the query results.
This is necessary because we require that the DAG only contains constant-normalized densities in order to sample from it (section \ref{sec:sampling}). \\
Discussed in section \ref{sec:querying}.
\item \emph{Generate sampling code.}
Generate a forward sampling program which draws each variable \(v\) from its constant-normalized conditional density function \(D_v\) determined by \(s\in S^*\).
Since this step is implementation specific, we discuss this in section \ref{sec:implementation} along with our Stan implementation.
\end{enumerate}

\begin{theorem}
\label{thm:uniqueness}
\noindent The edge selection sets \(s\) in \(S^*\) produce DAGs which are equal to each other up to proportionality.
\end{theorem}

\begin{proof}
Let \(s_1,\ s_2\in S^*\) and \(v \in V\).

Since \(s_1\) and \(s_2\) are both constant-normalized, \(D_{v, C(G_{Fac}, s_1)}(v)\) and \(D_{v, C(G_{Fac}, s_2)}(v)\) are both proportional to the marginal distribution for \(v\) according to Theorem \ref{thm:sampling}, and are therefore also proportional to each other.
\end{proof}

Due to Theorem \ref{thm:uniqueness}, any edge selection set \(s \in S^*\)  will result in the same sampling distribution.

\subsubsection{Recognizing constant-normalized conditional densities}
\label{sec:orgb1b3d1c}
\label{sec:recognizing}
Since we are unable to classify constant-normalization statically from source code in general, we use simple pattern heuristics to identify a conservative subset of cases which we call \emph{recognizable edges}.
When we recognize an edge \((v, f)\) of the factor graph, we assert that \(f\) makes up the entire conditional density for \(v\).

In our Stan implementation, we can recognize a factor \(f\) as a constant-normalized conditional density for \(v\) if the statement of \(f\) takes the form of \texttt{target += dist\(\_\)lpdf(v | ..)} or \texttt{v \(\sim\) dist(..)}, where \texttt{dist\(\_\)lpdf} is either one of Stan's built-in probability distributions\cite{stan-func-ref} or a user defined distribution (whose 
name ends in \texttt{\(\_\)lpdf}).
User-defined function names with this suffix are considered to be user annotations that assert that the function implements a constant-normalized distribution.

\subsubsection{Computing sound edge selection sets}
\label{sec:orgdfefc08}
\label{sec:computing}

Our strategy is to leverage a SAT solver into producing \(S\), the set of edge selection sets \(s\) such that \(\sigma(G_{Fac}, C(G_{Fac}, s))\).
SAT solvers take as input a propositional formula\footnote{The SAT solver interface we use allows us to input general propositional formulae, which are then automatically translated to Conjunctive Normal Form.\label{org059471a}} over a set of atomic propositions, and return the set of subsets of atomic propositions which satisfy the formula.
We will construct a propositional formula and set of atomic propositions to encode \(\sigma\) and then translate the set of solutions into the set \(S\).

First, we construct a set of atomic propositions to represent an edge selection set \(s\) and useful properties of \(s\).
The set of atomic propositions \(Atom(G_{Fac})\) is the union of the following:
\begin{enumerate}
\item \emph{An edge (v, f) is selected.}
\(Sel_{v, f}\) asserts that the edge \((v, f)\) is included in the selection set.
$$\{Sel_{v, f} \mid v \in V(G_{Fac}), f \in F(G_{Fac})\}$$
\item \emph{There exists a path between vertices.}
\(P_{v_1 \rightarrow v_2}\) asserts the existence of a path from \(v_1\) to \(v_2\) in \(C(G_{Fac}, s)\).
$$\{P_{v_1 \rightarrow v_2} \mid v_1,v_2 \in V(G_{Fac})\}$$
\end{enumerate}

Let \(\alpha : \mathcal{P}(Atom(G_{Fac})) \rightarrow \mathcal{P}(E(G_{Fac}))\) be a function that reconstructs an edge selection set from a SAT solution: \(\alpha(A) = \{(v, f)\mid Sel_{v, f} \in A\}\).

Next we construct a propositional formula, \(prop_{G_{Fac}}(A)\), to be \texttt{true} if and only if \(\sigma(G_{Fac}, C(G_{Fac}, \alpha(A)))\)
and \(r\subseteq \alpha(A)\).
The formula \(prop_{G_{Fac}}\) is the conjunction of the following rules:
\begin{enumerate}
\item \emph{The resulting DAG is acyclic.}
\(C(G_{Fac}, s)\) does not include any cycles, so no variable has a path to itself.
$$\forall v \in V(G_{Fac}), \neg P_{v\rightarrow v}$$
\item \emph{All factors are covered.}
\(s\) includes at least one edge that includes each factor in \(F(G_{Fac})\).
$$\forall f \in F(G_{Fac}), \exists v \in V(G_{Fac}),\ Sel_{v, f}$$
\item \emph{Factors are not covered more than once.}
\(s\) does not include more than one edge that includes each factor in \(F(G_{Fac})\).
Therefore, selecting one edge with a factor \(f\) implies that no other edges which include \(f\) are selected.
$$\forall f \in F(G_{Fac}), v_1 \in V(G_{Fac}), v_2 \in V(G_{Fac}),$$
$$(v_1 \neq v_2) \wedge Sel_{v_1, f} \implies \neg Sel_{v_2, f}$$
\item \emph{All variables are covered.}
\(s\) includes at least one edge that includes each variable in \(V(G_{Fac})\).
$$\forall v \in V(G_{Fac}), \exists f \in F(G_{Fac}),\ Sel_{v, f}$$
\item \emph{Edges in \(r\) are included.}
$$\forall (v, f) \in r,\ Sel_{v,f}$$
\item \emph{Variables covered by \(r\) are not included again.}
Since in a recognizable edge \((v, f)\) the factor \(f\) is asserted to be the entire conditional density for \(v\), no other factors will contribute to \(v\)'s conditional density.
$$\forall (v,f_1) \in r, f_2 \in F(G_{Fac}), f_1 \neq f_2 \implies \neg Sel_{v, f_2}$$
\item \emph{Selecting an edge creates edges in the DAG.}
There will be an edge from variable \(v_1\) to variable \(v_2\) in \(C(G_{Fac}, s)\) if, for some factor \(f\), there is an unselected edge \((v_1, f)\) (implying that \(v_1\) contributes to \(f\)) and a selected edge \((v_2, f)\).
$$\forall f \in F(G_{Fac}), v_1, v_2 \in V(G_{Fac}),$$
$$(v_1, f), (v_2, f) \in E(G_{Fac}),$$
$$\neg Sel_{v_1, f} \wedge Sel_{v_2, f} \implies P_{v_1 \rightarrow v_2}$$
\item \emph{Paths between variables compose.}
$$\forall v_1, v_2, v_3 \in V(G_{Fac}),\ P_{v_1 \rightarrow v_2} \wedge P_{v_2 \rightarrow v_3} \implies P_{v_1 \rightarrow v_3}$$
\end{enumerate}

We now construct the set \(S\) of \emph{sound edge selection sets} by:
$$ S = \{\alpha(A) \mid A \in \texttt{SAT}(prop_{G_{Fac}})\} $$
\begin{theorem}[Soundness]
\noindent For any sound edge selection set \(s\in S\), we have that \(r\subseteq s\) and  \(C(G_{Fac}, s)\) is a sound transformation of \(G_{Fac}\).
\end{theorem}
\begin{proof}
By rule 4 of \(prop_{G_{Fac}}\), each variable in \(G_{Fac}\) will appear in \(s\), and so \(V(G_{DAG}) = V(C(G_{Fac}, s))\).

By rules 2 and 3 of \(prop_{G_{Fac}}\), each factor will be included in exactly one edge in \(s\), so the joint density does not change:
\begin{align*}
pdf_{G_{DAG}}(V(C(G_{Fac}, s))) &= \prod_{v} \prod_{f\mid (v, f) \in s} f(Nei(G_{Fac}, f))\\
                      &= \prod_{f} f(Nei(G_{Fac}, f))\\
                      &= pdf_{G_{Fac}}(V(G_{Fac}))
\end{align*}
\end{proof}
\begin{theorem}[Completeness]
\noindent For any edge selection set \(s\) such that \(C(G_{Fac}, s)\) is a sound transformation of \(G_{Fac}\) and \(r\subseteq s\),
\(prop_{G_{Fac}}(s)\) is \texttt{true}.
\end{theorem}
\begin{proof}
Suppose that there were some \(s=\alpha(A)\) for which \(G=C(G_{Fac}, s)\) is a sound transformation of \(G_{Fac}\), but \(prop_{G_{Fac}}(A)\) were \texttt{false}.
\(prop_{G_{Fac}}(A)\) being \texttt{false} implies that \(s\) broke at least one rule:

If \(s\) broke rule 1, \(G\) would be cyclic and therefore not a DAG.

If \(s\) broke rule 2, \(G\) would not include some factor in its joint density function and would not have the same joint density as \(G_{Fac}\), and so would not be sound.

If \(s\) broke rule 3, \(G\) would include some factor in its joint density function more than once and would not have the same joint density as \(G_{Fac}\), and so would not be sound.

If \(s\) broke rule 4, \(G\) would not include some variable in \(V(G_{DAG})\), and so would not be sound.

If \(s\) broke rule 5, \(r\not\subseteq s\).

If \(s\) broke rule 6, some \(v\) covered by \(r\) would have another edge, so the variables covered by \(r\) would otherwise assigned.

Rules 7 and 8 enforce the invariant that the \(P_{\_ \rightarrow \_}\) propositions reflect the paths in the DAG produced by the \(Sel\) variables, and they cannot in themselves reject any potential \(s\).

Breaking any of these rules results in a contradiction, so no such \(s\) exists.
\end{proof}
\noindent Consequently, \(prop_{G_{Fac}}(A) \Leftrightarrow \sigma(G_{DAG}, C(G_{DAG}, \alpha(s)))\wedge r\subseteq s\). Therefore, \(\{s(A)\mid A \in \texttt{SAT}(prop_{G_{Fac}})\} = \{s \mid \sigma(G_{DAG}, C(G_{DAG}, s))\}\) is a sound and complete set of edge selection sets.

\subsubsection{Querying the user with ambiguous conditional densities}
\label{sec:orgac2827a}
\label{sec:querying}
When it cannot be gleaned whether a conditional density is constant-normalized, either from a recognizable edge or from the topology of the factor graph, then it is necessary to ask the user if they can assert constant-normalization.
This is necessary because it is not possible to perfectly identify constant-normalization from code, while we require all densities to be constant-normalized in order to sample from the DAG (section \ref{sec:sampling}).

There are alternatives to avoid user queries:
(1) Wrap each constant-normalized conditional density function in a recognizable user-defined function before the tool is applied.
This will disambiguate all conditional densities.
(2) Fail when conditional densities cannot be automatically proven constant-normalized.
This will limit the scope of application of the tool.

We describe our strategy for user queries below.

Given the set \(S\) of sound edge selection sets and the set \(r\) of recognizable edges, our goal is to filter out edge selection sets in \(S\) which contain conditional density assignments that are not constant-normalized.

We will query the user with each density assignment, allowing them to assert or not assert that the presented set of factors is constant-normalized.
We will not need to query assignments which are included in \(r\) since those are already assumed to be constant-normalized.
We will also not need to query any variables which are root variables in all DAGs arising from \(S\).
We can then filter out edge selection sets with assignments which the user did not assert are constant-normalized.

We use a query procedure, \texttt{query}, which takes as input a variable \(v\) and a set of assigned factors \(F\) and returns \texttt{true} if and only if the user asserts that the assigned density \(D_v(v \mid \dots) = \prod_{f\in F} f(Nei(G_{Fac}, f))\) is constant-normalized.

We construct the set of queries \(Q\):
\begin{align*}
&V' = V \setminus \{v \mid (v, f) \in r\}\\
&F(v, s) = \{f \mid (v, f) \in s\}\\
&Q = \{(v, F(v, s)) \mid v \in V',\ s \in S \}
\end{align*}

The set of affirmative queries \(Q'\) is:
$$Q' = \{(v, F) \mid (v, F) \in Q\,\ \texttt{query}(v, F) \}$$

We can then produce the set of edge selection sets which are consistent with the affirmative queries, \(S^*\):
$$S^* = \{ s \mid s \in S\,\ \forall v\in V',\ (v, F(v, s)) \in Q'\}$$

When the set \(S^*\) is empty, we can conclude that no DAG can be derived from the factor graph with the given constant-normalized densities.

\section{Results}
\label{sec:org53607fe}
\subsection{Stan implementation}
\label{sec:org8d92e1b}
\label{sec:implementation}
We implemented a pipeline that
(1) extracts a factor graph from a Stan program,
(2) produces a constant-normalized DAG\footnote{Since multiple valid DAGs will have proportional conditional densities (Theorem \ref{thm:uniqueness}), it is not necessary to return the whole set.\label{orgdb435b8}} when possible using a SAT solver and possibly user queries, and then
(3) constructs a forward-sampling program.
Parts (1) and (3) are specific to Stan (but could be implemented for any Stan-like language), while part (2) could readily be applied to a factor graph from any source\footnote{The only Stan-specific part of the DAG transformation is the heuristic to recognize constant-normalized edges; this would need to be adapted in an implementation for another language.\label{org3a81814}}.

\subsubsection{Extracting a factor graph from a Stan program}
\label{sec:org6e3137c}
Factor graphs are derived from Stan programs by a pipeline of static analyses which we have built into the open-source Stanc3 compiler \cite{stanc3}.

Consider a Stan program \(P\) with statement set \(S(P)\) and a set of free variables \(V(P)\).
Let \(V_S(S)\) be the set of free variables in the statement \(S\).

We first derive a dependency graph \(D(P)\) which is a directed graph between the statements \(S(P)\). 
The dependency graph has a directed edge \((s_1, s_2) \in E(D(P))\) if \(s_1\) directly or indirectly influences the behavior of \(s_2\).
The dependency graph is built using a reaching definitions monotone framework analysis on the control flow graph  \cite{nielson2015principles}.
As is standard, the resulting static analysis is sound but not complete.
As such it will detect a superset of the true dependencies present in the program, resulting in a conservative approximation.

From the dependency graph \(D(P)\) and a statement \(S\), we construct a set \(V_D(S)\) to be the set of free variables that \(S\) depends on:
$$V_D(s_1) = V_S(s_1) \cup \{ v \mid (s_2, s_1) \in E(D(P)),\ v \in V_S(s_2)\}$$

Next, we collect all factors in the program, \(F(P)\).
Factors are statements which have a direct effect on the density that the Stan program represents.
These take the form of:
\begin{enumerate}
\item \texttt{target += ..} and \texttt{x \(\sim\) ..} statements, which directly increment the \texttt{target} value.
\item Function calls which can affect \texttt{target}.
In Stan, only functions which the suffix \texttt{\_lp} can affect target.
\item \texttt{reject} statements.
In Stan, when a \texttt{reject} statement is executed, the current sample point is thrown out, which implicitly sets the density of the point to zero.
\end{enumerate}
For each statement \(s\) of the above form, we decorate \(s\) with its statement dependencies to produce each factor \(f \in F(P)\):
$$f = (s, \{s' \mid (s', s) \in E(D(P)) \})$$
We can now construct a factor graph \(G_{Fac}(P)\):
\begin{align*}
V(G_{Fac}(P)) &= V(P)\\
F(G_{Fac}(P)) &= F(P)\\
E(G_{Fac}(P)) &= \{(v, f)\mid f\in F(P),\ v\in V_D(f))
\end{align*}

\subsubsection{Producing a constant-normalized DAG}
\label{sec:org44d1ad5}
We implement the algorithm described in section \ref{sec:computing} with a simple Haskell program using the MiniSat package \cite{sorensson2005minisat,haskell-miniset-bindings,implementation}.
The implementation parses the factor graph produced by the Stanc3 compiler, applies a SAT solver, and queries the user when the constant-normalization of a density is ambiguous (section \ref{sec:querying}).

If this process returns no solutions, we can conclude that no sound DAG can be produced from the original factor graph with the given constant-normalized densities, and our process terminates.

\subsubsection{Generating sampling code}
\label{sec:orgfd9c4ae}
\label{sec:translate-factors}
Finally, we construct a sampling program using the DAG \(G\) and the program statements which decorate the factors.
For a set $A$ of factors, let \(Stat(A)\) be the sequence of statements in $P$ corresponding to the factors
and write
\(Slice(A)\) for their sequence of statement dependencies in \(P\).

We first define \(\texttt{sample}(A)\) which generates the necessary code for \(A\), the set of factors assigned to the conditional density function.
\(\texttt{sample}(A)\) returns a list of statements which either generate samples with a Stan builtin \texttt{\_rng} function and are decorated with an \texttt{RNG} label, or calculate a density function by affecting Stan's \texttt{target} variable and are decorated with a \texttt{PDF} label:
\begin{align*}
\texttt{samp}&\texttt{le}(A) =\\
  \texttt{if}&\text{ $A$ is a singleton factor $\{f\}$ with $f$ of the form:} \\
   &\ \ \texttt{target += dist\_lpdf(mu | ..)}\quad \textnormal{or}\\
   & \ \  \texttt{mu $\sim$ dist(..)}\\
  \texttt{th}&\texttt{en} \\
   &\ \ \big(\texttt{RNG},\ Slice(\{f\}) +\!\!\!+\ \texttt{[mu = dist\_rng(..)]} \big)  \\
  \texttt{el}&\texttt{se} \\
   &\ \ \big(\texttt{PDF},\  Slice(A)+\!\!\!+\ Stat(A)\big)
\end{align*}

We can now apply the procedure \texttt{sampleG} defined in section \ref{sec:sampling}, which applies \texttt{sample} to each variable's assigned factors in topological order and returns a sequence of the results.
The result is a sequence of code segments decorated with \texttt{RNG} or \texttt{PDF}.

\subsubsection{Synthesizing sampling programs as Stan programs}
\label{sec:orgfb113b4}
The way that we can turn the sequence of code segments into executable Stan programs depends on the number of segments labeled as \texttt{PDF}.
These statements can only be written in Stan's \texttt{model} block, while segments labeled as \texttt{RNG} can only be written in Stan's \texttt{transformed data} or \texttt{generated quantities} blocks, which are executed before and after \texttt{model}, respectively.
Since Stan programs only have one \texttt{model} block, we will need one Stan program for each \texttt{PDF}-labeled segment.

For example, suppose we have the following sequence of code segments:
$$(\texttt{RNG}, S_1), (\texttt{RNG}, S_2), (\texttt{PDF}, S_3), (\texttt{RNG}, S_4)$$
\(S_3\) must go in the \texttt{model} block.
Since \(S_1\) and \(S_2\) must go before \(S_3\), they go in the \texttt{transformed data} block.
Since \(S_4\) must go after \(S_3\), it goes in the \texttt{generated quantities} block.

\subsubsection{Synthesizing Stan programs for prior predictive sampling}
\label{sec:orgc90c455}
\label{sec:prior-pred}
Prior predictive sampling is done by sampling from the model parameters without influence from the data, and then sampling synthetic data from the likelihood, using the model parameter draws.

We generate the sequence \(Sam_{prior}\) of sampling code for the prior samples by removing each data variable and its neighbors from the factor graph, and then following our method.
We then generate the sequence \(Sam_{predictive}\) of sampling code for the data variables by removing each model parameter variables from the factor graph, and then following our method.
We then synthesize a Stan program from \(Sam_{prior} +\!\!\!+\ Sam_{predictive}\).
\subsubsection{Synthesizing Stan programs for Simulation-based Calibration}
\label{sec:org09b3159}
Simulation-based Calibration is performed by sampling from the prior predictive distribution, then sampling from the posterior distribution given the synthetic data, and finally calculating the rank statistics of the generating prior samples among the posterior samples.
This procedure can be achieved within Stan by using the \texttt{model} block to draw from the posterior and the \texttt{generated quantities} block to calculate the rank statistics \cite{users-guide-sbc}.
The prior predictive sampling code is generated in the same way as section \ref{sec:prior-pred}.
Because SBC requires the use of a \texttt{model} block to sample from the posterior, the whole process cannot be fit into one Stan program unless there are no \texttt{PDF} code segments in the prior predictive sampling code. Otherwise it will need one Stan program for each \texttt{PDF}-labeled segment plus one.
\subsection{Stan program example}
\label{sec:orge5dfd43}
We will step through two examples of translating Stan programs into forward sampling form.
The first is a simple example where most edges are recognizable, no user queries are necessary, and a single-program SBC can be synthesized.
The second is a more complex example where most edges are not recognizable and a user query is necessary.
\subsubsection{Eight schools example}
\label{sec:orga369664}
Suppose we have the following Stan program, which is modified\footnote{The non-distribution expression on line 12 was added to make an instructive example, and the distribution on line 13 was added to give \texttt{tau} an arbitrary prior.\label{org0fa3555}} from the well-known eight schools problem \cite{BDA}:

\VerbatimInputNumbered{examples/eight_schools.stan}

Our goal will be to produce a new Stan program which draws from the prior predictive distribution for \texttt{y} \cite{gabry2019visualization}.

The Stanc3 compiler emits the following factor graph \(G_{Fac}\):
\begin{center}
\includegraphics[width=.9\linewidth]{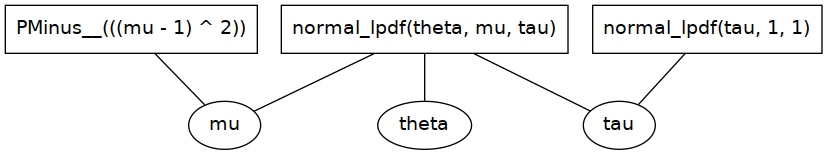}
\end{center}

To translate this factor graph to a DAG, our implementation follows the steps from \ref{sec:computing}:

\begin{enumerate}
\item \emph{Identify the set of recognizable edges, \(r\).}
The recognizable set is \(r=\{(\texttt{theta},\ \texttt{normal\_lpdf(theta, mu, tau)})\),\\ \((\texttt{tau},\ \texttt{normal\_lpdf(tau, 1, 1)})\}\).
\item \emph{Construct the set of valid edge selection sets, \(S\).}
Given the recognizable edges, the remaining variables to be covered are \texttt{mu} and \texttt{tau}.
Each only has one option, so \(S=r \cup \{(\texttt{mu},\ \texttt{-(mu - 1)\UseVerb{carot}2})\}\).
\item \emph{Query the user when necessary.}
We do not need to query the user in this case, because all variables either have recognizable edges (\texttt{theta} and \texttt{tau}) or are root variables in all DAGs arising from \(S\) and are therefore trivially constant-normalized (\texttt{mu} and \texttt{tau}).
Since we are now confident that each assigned density in \(S\) is constant-normalized, we can assert that the element \(s\) of \(S\) will produce a constant-normalized DAG \(G_{DAG}=C(G_{Fac}, s)\):
\begin{center}
\includegraphics[width=1.0in]{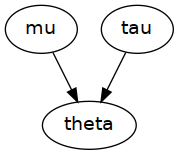}
\end{center}
\item \emph{Generate sampling code.}
We first transform and decorate each factor as in section \ref{sec:translate-factors}, producing the sequence \(SAM_{prior}\):
\begin{align*}
&(\texttt{PDF},\ \texttt{target += -(mu - 1)\UseVerb{carot}2})\\
&(\texttt{RNG},\ \texttt{tau = normal\_rng(1, 1)})\\
&(\texttt{RNG},\ \texttt{theta = normal\_rng(mu, tau)})
\end{align*}
\end{enumerate}

To produce PPC  code as in section \ref{sec:prior-pred}, we repeat the above process starting with a factor graph that includes the desired data variables as vertices and holds the other data variables and model parameter variables constant.
We find that this second DAG is built from one recognizable edge \((\texttt{y},\ \texttt{normal\_lpdf(y, theta, sigma)})\).
We can transform and decorate this factor to produce the sequence \(SAM_{predictive}\):
\begin{align*}
&(\texttt{RNG},\ \texttt{y = normal\_rng(theta, sigma)})
\end{align*}

We now generate code for \(SAM_{prior} +\!\!\!+\ SAM_{predictive}\).
Since there is only one \texttt{PDF} element, we can produce prior predictive sampling code in a single Stan program:
\VerbatimInputNumbered{examples/eight_schools.stan.ppc.stan}
This was a simple example where we produce a sampling program automatically.
\subsubsection{Query example}
\label{sec:org206ffe9}
\label{sec:query-example}
Next consider the following Stan program:

\VerbatimInputNumbered{examples/ex1.stan}

Our goal will be to produce a constant-normalized DAG without the data variables, which could then produce a sampling program.
Each line from 12 to 16 represents a non-data factor. We will refer to these factors by line number, e.g., \(F_{12}=\texttt{normal\_lpdf(b | 1, e)}\).

The factors \(F_{15}\) and \(F_{16}\) together make up an inverse Gaussian distribution on \texttt{e} with mean \texttt{c} and shape parameter \texttt{d}.

The Stanc3 compiler emits the following factor graph \(G_{Fac}\) (after renaming):
\begin{center}
\includegraphics[width=2.7in]{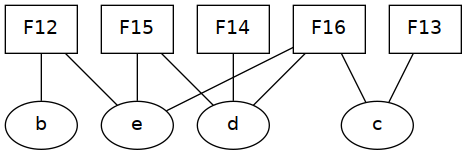}
\end{center}
Again we follow the steps from section \ref{sec:computing}:

\begin{enumerate}
\item \emph{Identify the set of recognizable edges, \(r\).}
The recognizable set is \(r=\{(\texttt{b},\ F_{12})\}\).
\item \emph{Construct the set of valid edge selection sets, \(S\).}
Our implementation finds two valid edge selection sets, one in which \(F_{16}\) is matched with \texttt{e} and one in which it is matched with \texttt{c}.
All other ambiguity is eliminated automatically: for example, matching \(F_{16}\) or \(F_{15}\) with \texttt{d} would imply a cycle of \texttt{d} and \texttt{e}.
\item \emph{Query the user when necessary.}
To disambiguate between the two elements of \(S\), it is necessary to query the user to determine the true constant-normalized distributions for either \texttt{d} or \texttt{e} (the distribution for one is sufficient to infer the other).
Out implementation prompts:
\begin{Verbatim}
Which of the following is a constant-normalized
density for e?
  0: None of the below
  1: lpdf(e | c, d) = 
        ((PMinus__(d) * ((e - c) ^ 2))
                           / ((2 * (c ^ 2)) * e))
      + (0.5 * log((((d / 2) * pi()) * (e ^ 3))))
  2: lpdf(e | d) = 
        (0.5 * log((((d / 2) * pi()) * (e ^ 3))))
Enter a number 0-2: 
\end{Verbatim}
If the user selects \texttt{0}, no DAG will be produced; if the user selects \texttt{1}, the correct DAG will be produced (\textbf{(a)} below); if the user selects \texttt{2}, an alternate DAG will be produced (\textbf{(b)} below).


\begin{figure}[htpb]
  \centering
  \subcaptionbox{}{%
    \includegraphics[width=1.0in]{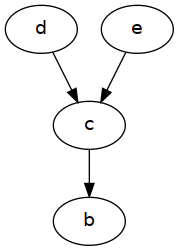}
    \label{fig:sub1}
  }
  \subcaptionbox{}{%
    \includegraphics[width=1.0in]{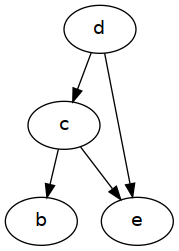}
    \label{fig:sub2}
  }
\end{figure}


\end{enumerate}

\section{Discussion}
\label{sec:org1e96897}
One limitation of our approach is that we rely on a SAT solver to perform the graph transformation between factor graphs and DAGs.
Since the SAT problem is NP-Hard, we cannot provide a polynomial computational bound on the graph transformation nor size bound on the number of edge selection sets, so sufficiently large or pathological programs will not be practical applications of this approach.

An additional limitation is that we rely on static analysis to derive the dependency structure of Stan programs.
This static analysis has a number of limitations,
all of which can jeopardize the efficiency but not the correctness of our approach:
\begin{itemize}
\item Container types such as arrays are treated as monolithic variables, which potentially overestimates dependence.
This will be mitigated by further engineering effort on the Stan compiler, for example by leveraging well-known techniques like polyhedral dependence analysis \cite{feautrier1992some}.
\item Control flow and other value-dependent structure cannot be perfectly predicted due to the halting problem, which can hide dependence structure.
\item There is independence structure inside some Stan math distributions which is not accounted for.
For example, a multivariate Gaussian distribution with diagonal covariance matrix does not introduce dependency between each dimension of its variate.
\end{itemize}

Our method sometimes requires user interaction (avoiding user interaction was discussed in section \ref{sec:querying}).
Since the user interaction is minimal and avoidable, we consider this a small limitation.
We expect that users will typically find the answer to a query to be obvious, since the code was likely produced with a probability distribution in mind for each variable, as in section \ref{sec:query-example}.
If the user is unsure, they can simply make no claims of constant-normalization and result will be a loss of efficiency rather than loss of correctness.
Only in cases where the user makes an incorrect assertion will the correctness of the result be compromised, and that case a hand-built sampling program would certainly contain the same error.

We do not believe that these limitations will prevent this method from being applicable in the vast majority of practical cases.
We hope that this tool will fit into the workflow of many Stan users, allowing them to efficiently automate methods like SBC and PPC and thereby produce more reliable results.

We also note that although parts of our implementation are specific to Stan, the idea is applicable more generally to Stan-like density-based probabilistic programming languages \cite{the-tome}.
There may also be other applications for the production of DAGs from Stan-like programs. For example, for directly checking the structure of the written model.
  When a Stan user writes a Stan program, they will sometimes have expectations about the independence structure in terms of a directed graph, since directed probabilistic graphical models are common representations in machine learning \cite{bayes-nets} and effective at communicating modeling ideas.
  When this is the case, we could automatically check the user's Stan program against their assumptions by providing a visualization of the DAG, validating the program against a representation of the DAG, or verifying certain graphical properties of the program like its hierarchical structure.

\section{Acknowledgments}
Matthijs V\'ak\'ar was funded by the European Union’s Horizon 2020 research and innovation programme under the Marie Skłodowska-Curie grant agreement No. 895827.
Ryan Bernstein and Jeannette Wing supported by the Columbia University Data Science Institute.
Ryan Bernstein was also supported by the National Science Foundation under grant number 1730414
and the Office of Naval Research under award number N00014-19-1-2204,
as well as generous research awards from Facebook through the Probability and Programming 2020 program and from the Schmidt Foundation.
This work benefited from discussions with Bob Carpenter, Andrew Gelman, Sean Talts.

\bibliography{acmart}


\begin{thebibliography}{20}


\ifx \showCODEN    \undefined \def \showCODEN     #1{\unskip}     \fi
\ifx \showDOI      \undefined \def \showDOI       #1{#1}\fi
\ifx \showISBNx    \undefined \def \showISBNx     #1{\unskip}     \fi
\ifx \showISBNxiii \undefined \def \showISBNxiii  #1{\unskip}     \fi
\ifx \showISSN     \undefined \def \showISSN      #1{\unskip}     \fi
\ifx \showLCCN     \undefined \def \showLCCN      #1{\unskip}     \fi
\ifx \shownote     \undefined \def \shownote      #1{#1}          \fi
\ifx \showarticletitle \undefined \def \showarticletitle #1{#1}   \fi
\ifx \showURL      \undefined \def \showURL       {\relax}        \fi
\providecommand\bibfield[2]{#2}
\providecommand\bibinfo[2]{#2}
\providecommand\natexlab[1]{#1}
\providecommand\showeprint[2][]{arXiv:#2}

\bibitem[\protect\citeauthoryear{Becker, Huang, Bieri, Ma, Knowles,
  Jafar-Nejad, Messing, Kim, Soriano, Auburger, et~al\mbox{.}}{Becker
  et~al\mbox{.}}{2017}]%
        {becker2017therapeutic}
\bibfield{author}{\bibinfo{person}{Lindsay~A Becker}, \bibinfo{person}{Brenda
  Huang}, \bibinfo{person}{Gregor Bieri}, \bibinfo{person}{Rosanna Ma},
  \bibinfo{person}{David~A Knowles}, \bibinfo{person}{Paymaan Jafar-Nejad},
  \bibinfo{person}{James Messing}, \bibinfo{person}{Hong~Joo Kim},
  \bibinfo{person}{Armand Soriano}, \bibinfo{person}{Georg Auburger},
  {et~al\mbox{.}}} \bibinfo{year}{2017}\natexlab{}.
\newblock \showarticletitle{Therapeutic reduction of ataxin-2 extends lifespan
  and reduces pathology in TDP-43 mice}.
\newblock \bibinfo{journal}{\emph{Nature}} \bibinfo{volume}{544},
  \bibinfo{number}{7650} (\bibinfo{year}{2017}), \bibinfo{pages}{367--371}.
\newblock


\bibitem[\protect\citeauthoryear{Bernstein}{Bernstein}{2019}]%
        {the-tome}
\bibfield{author}{\bibinfo{person}{Ryan Bernstein}.}
  \bibinfo{year}{2019}\natexlab{}.
\newblock \showarticletitle{Static Analysis for Probabilistic Programs}.
\newblock \bibinfo{journal}{\emph{arXiv preprint arXiv:1909.05076}}
  (\bibinfo{year}{2019}).
\newblock


\bibitem[\protect\citeauthoryear{Bernstein}{Bernstein}{2020}]%
        {implementation}
\bibfield{author}{\bibinfo{person}{Ryan Bernstein}.}
  \bibinfo{year}{2020}\natexlab{}.
\newblock \bibinfo{title}{rybern/factor-graph-to-dag}.
\newblock
\newblock
\urldef\tempurl%
\url{https://github.com/rybern/factor-graph-to-dag}
\showURL{%
\tempurl}


\bibitem[\protect\citeauthoryear{Burnett, Chen, Szyszkowicz, Fann, Hubbell,
  Pope, Apte, Brauer, Cohen, Weichenthal, et~al\mbox{.}}{Burnett
  et~al\mbox{.}}{2018}]%
        {burnett2018global}
\bibfield{author}{\bibinfo{person}{Richard Burnett}, \bibinfo{person}{Hong
  Chen}, \bibinfo{person}{Mieczys{\l}aw Szyszkowicz}, \bibinfo{person}{Neal
  Fann}, \bibinfo{person}{Bryan Hubbell}, \bibinfo{person}{C~Arden Pope},
  \bibinfo{person}{Joshua~S Apte}, \bibinfo{person}{Michael Brauer},
  \bibinfo{person}{Aaron Cohen}, \bibinfo{person}{Scott Weichenthal},
  {et~al\mbox{.}}} \bibinfo{year}{2018}\natexlab{}.
\newblock \showarticletitle{Global estimates of mortality associated with
  long-term exposure to outdoor fine particulate matter}.
\newblock \bibinfo{journal}{\emph{Proceedings of the National Academy of
  Sciences}} \bibinfo{volume}{115}, \bibinfo{number}{38}
  (\bibinfo{year}{2018}), \bibinfo{pages}{9592--9597}.
\newblock


\bibitem[\protect\citeauthoryear{Carpenter, Gelman, Hoffman, Lee, Goodrich,
  Betancourt, Brubaker, Guo, Li, and Riddell}{Carpenter et~al\mbox{.}}{2017}]%
        {carpenter2017stan}
\bibfield{author}{\bibinfo{person}{Bob Carpenter}, \bibinfo{person}{Andrew
  Gelman}, \bibinfo{person}{Matthew~D Hoffman}, \bibinfo{person}{Daniel Lee},
  \bibinfo{person}{Ben Goodrich}, \bibinfo{person}{Michael Betancourt},
  \bibinfo{person}{Marcus Brubaker}, \bibinfo{person}{Jiqiang Guo},
  \bibinfo{person}{Peter Li}, {and} \bibinfo{person}{Allen Riddell}.}
  \bibinfo{year}{2017}\natexlab{}.
\newblock \showarticletitle{Stan: A probabilistic programming language}.
\newblock \bibinfo{journal}{\emph{Journal of statistical software}}
  \bibinfo{volume}{76}, \bibinfo{number}{1} (\bibinfo{year}{2017}).
\newblock


\bibitem[\protect\citeauthoryear{Feautrier}{Feautrier}{1992}]%
        {feautrier1992some}
\bibfield{author}{\bibinfo{person}{Paul Feautrier}.}
  \bibinfo{year}{1992}\natexlab{}.
\newblock \showarticletitle{Some efficient solutions to the affine scheduling
  problem. I. One-dimensional time}.
\newblock \bibinfo{journal}{\emph{International journal of parallel
  programming}} \bibinfo{volume}{21}, \bibinfo{number}{5}
  (\bibinfo{year}{1992}), \bibinfo{pages}{313--347}.
\newblock


\bibitem[\protect\citeauthoryear{Gabry, Simpson, Vehtari, Betancourt, and
  Gelman}{Gabry et~al\mbox{.}}{2019}]%
        {gabry2019visualization}
\bibfield{author}{\bibinfo{person}{Jonah Gabry}, \bibinfo{person}{Daniel
  Simpson}, \bibinfo{person}{Aki Vehtari}, \bibinfo{person}{Michael
  Betancourt}, {and} \bibinfo{person}{Andrew Gelman}.}
  \bibinfo{year}{2019}\natexlab{}.
\newblock \showarticletitle{Visualization in Bayesian workflow}.
\newblock \bibinfo{journal}{\emph{Journal of the Royal Statistical Society:
  Series A (Statistics in Society)}} \bibinfo{volume}{182}, \bibinfo{number}{2}
  (\bibinfo{year}{2019}), \bibinfo{pages}{389--402}.
\newblock


\bibitem[\protect\citeauthoryear{Gelman, Carlin, Stern, Dunson, Vehtari, and
  Rubin}{Gelman et~al\mbox{.}}{2013}]%
        {BDA}
\bibfield{author}{\bibinfo{person}{Andrew Gelman}, \bibinfo{person}{John~B
  Carlin}, \bibinfo{person}{Hal~S Stern}, \bibinfo{person}{David~B Dunson},
  \bibinfo{person}{Aki Vehtari}, {and} \bibinfo{person}{Donald~B Rubin}.}
  \bibinfo{year}{2013}\natexlab{}.
\newblock \bibinfo{booktitle}{\emph{Bayesian data analysis}}.
\newblock \bibinfo{publisher}{CRC press}.
\newblock


\bibitem[\protect\citeauthoryear{Geyer}{Geyer}{1992}]%
        {MCMC}
\bibfield{author}{\bibinfo{person}{Charles~J Geyer}.}
  \bibinfo{year}{1992}\natexlab{}.
\newblock \showarticletitle{Practical markov chain monte carlo}.
\newblock \bibinfo{journal}{\emph{Statistical science}} (\bibinfo{year}{1992}),
  \bibinfo{pages}{473--483}.
\newblock


\bibitem[\protect\citeauthoryear{Jensen et~al\mbox{.}}{Jensen
  et~al\mbox{.}}{1996}]%
        {bayes-nets}
\bibfield{author}{\bibinfo{person}{Finn~V Jensen} {et~al\mbox{.}}}
  \bibinfo{year}{1996}\natexlab{}.
\newblock \bibinfo{booktitle}{\emph{An introduction to Bayesian networks}}.
  Vol.~\bibinfo{volume}{210}.
\newblock \bibinfo{publisher}{UCL press London}.
\newblock


\bibitem[\protect\citeauthoryear{Kennedy, Simpson, and Gelman}{Kennedy
  et~al\mbox{.}}{2019}]%
        {kennedy2019experiment}
\bibfield{author}{\bibinfo{person}{Lauren Kennedy}, \bibinfo{person}{Daniel
  Simpson}, {and} \bibinfo{person}{Andrew Gelman}.}
  \bibinfo{year}{2019}\natexlab{}.
\newblock \showarticletitle{The Experiment is just as Important as the
  Likelihood in Understanding the Prior: a Cautionary Note on Robust Cognitive
  Modeling}.
\newblock \bibinfo{journal}{\emph{Computational Brain \& Behavior}}
  \bibinfo{volume}{2}, \bibinfo{number}{3-4} (\bibinfo{year}{2019}),
  \bibinfo{pages}{210--217}.
\newblock


\bibitem[\protect\citeauthoryear{Lieu, Farr, Betancourt, Smith, Sereno, and
  McCarthy}{Lieu et~al\mbox{.}}{2017}]%
        {lieu2017hierarchical}
\bibfield{author}{\bibinfo{person}{Maggie Lieu}, \bibinfo{person}{Will~M Farr},
  \bibinfo{person}{Michael Betancourt}, \bibinfo{person}{Graham~P Smith},
  \bibinfo{person}{Mauro Sereno}, {and} \bibinfo{person}{Ian~G McCarthy}.}
  \bibinfo{year}{2017}\natexlab{}.
\newblock \showarticletitle{Hierarchical inference of the relationship between
  Concentration and Mass in Galaxy Groups and Clusters}.
\newblock \bibinfo{journal}{\emph{Monthly Notices of the Royal Astronomical
  Society}} \bibinfo{volume}{468}, \bibinfo{number}{4} (\bibinfo{year}{2017}),
  \bibinfo{pages}{4872--4886}.
\newblock


\bibitem[\protect\citeauthoryear{Nielson, Nielson, and Hankin}{Nielson
  et~al\mbox{.}}{2015}]%
        {nielson2015principles}
\bibfield{author}{\bibinfo{person}{Flemming Nielson}, \bibinfo{person}{Hanne~R
  Nielson}, {and} \bibinfo{person}{Chris Hankin}.}
  \bibinfo{year}{2015}\natexlab{}.
\newblock \bibinfo{booktitle}{\emph{Principles of program analysis}}.
\newblock \bibinfo{publisher}{Springer}.
\newblock


\bibitem[\protect\citeauthoryear{Selinger}{Selinger}{2016}]%
        {haskell-miniset-bindings}
\bibfield{author}{\bibinfo{person}{Peter Selinger}.}
  \bibinfo{year}{2016}\natexlab{}.
\newblock \bibinfo{title}{minisat-solver: High-level Haskell bindings for the
  MiniSat SAT solver}.
\newblock
\newblock
\urldef\tempurl%
\url{https://hackage.haskell.org/package/minisat-solver}
\showURL{%
\tempurl}


\bibitem[\protect\citeauthoryear{Sorensson and Een}{Sorensson and Een}{2005}]%
        {sorensson2005minisat}
\bibfield{author}{\bibinfo{person}{Niklas Sorensson} {and}
  \bibinfo{person}{Niklas Een}.} \bibinfo{year}{2005}\natexlab{}.
\newblock \showarticletitle{Minisat v1. 13-a sat solver with conflict-clause
  minimization}.
\newblock \bibinfo{journal}{\emph{SAT}} \bibinfo{volume}{2005},
  \bibinfo{number}{53} (\bibinfo{year}{2005}), \bibinfo{pages}{1--2}.
\newblock


\bibitem[\protect\citeauthoryear{stan dev}{stan dev}{2020}]%
        {stanc3}
\bibfield{author}{\bibinfo{person}{stan dev}.} \bibinfo{year}{2020}\natexlab{}.
\newblock \bibinfo{title}{A New Stan-to-C++ Compiler, stanc3}.
\newblock
\newblock
\urldef\tempurl%
\url{https://github.com/stan-dev/stanc3}
\showURL{%
\tempurl}


\bibitem[\protect\citeauthoryear{Talts, Betancourt, Simpson, Vehtari, and
  Gelman}{Talts et~al\mbox{.}}{2018}]%
        {SBC}
\bibfield{author}{\bibinfo{person}{Sean Talts}, \bibinfo{person}{Michael
  Betancourt}, \bibinfo{person}{Daniel Simpson}, \bibinfo{person}{Aki Vehtari},
  {and} \bibinfo{person}{Andrew Gelman}.} \bibinfo{year}{2018}\natexlab{}.
\newblock \showarticletitle{Validating Bayesian Inference Algorithms with
  Simulation-Based Calibration}.
\newblock \bibinfo{journal}{\emph{arXiv: Methodology}} (\bibinfo{year}{2018}).
\newblock


\bibitem[\protect\citeauthoryear{Taylor and Letham}{Taylor and Letham}{2018}]%
        {taylor2018forecasting}
\bibfield{author}{\bibinfo{person}{Sean~J Taylor} {and}
  \bibinfo{person}{Benjamin Letham}.} \bibinfo{year}{2018}\natexlab{}.
\newblock \showarticletitle{Forecasting at scale}.
\newblock \bibinfo{journal}{\emph{The American Statistician}}
  \bibinfo{volume}{72}, \bibinfo{number}{1} (\bibinfo{year}{2018}),
  \bibinfo{pages}{37--45}.
\newblock


\bibitem[\protect\citeauthoryear{Team}{Team}{2020a}]%
        {stan-func-ref}
\bibfield{author}{\bibinfo{person}{Stan~Development Team}.}
  \bibinfo{year}{2020}\natexlab{a}.
\newblock \bibinfo{title}{Stan Functions Reference}.
\newblock
\newblock
\urldef\tempurl%
\url{https://mc-stan.org/docs/2_21/functions-reference/}
\showURL{%
\tempurl}


\bibitem[\protect\citeauthoryear{Team}{Team}{2020b}]%
        {users-guide-sbc}
\bibfield{author}{\bibinfo{person}{Stan~Development Team}.}
  \bibinfo{year}{2020}\natexlab{b}.
\newblock \bibinfo{title}{Stan User’s Guide}.
\newblock
\newblock
\urldef\tempurl%
\url{https://mc-stan.org/docs/2_23/stan-users-guide/sbc-in-stan.html}
\showURL{%
\tempurl}


\end{thebibliography}
\end{document}